# The Role of Calculi in Uncertain Reasoning

Michael P. Wellman*  
MIT

David E. Heckerman†  
Stanford University

Much of the controversy about methods for automated decision making has focused on specific calculi for combining beliefs or propagating uncertainty. We broaden the debate by (1) exploring the constellation of secondary tasks surrounding any primary decision problem, and (2) identifying knowledge engineering concerns that present additional representational tradeoffs. We argue on pragmatic grounds that the attempt to support all of these tasks within a single calculus is misguided. In the process, we note several uncertain reasoning objectives that conflict with the Bayesian ideal of complete specification of probabilities and utilities. In response, we advocate treating the uncertainty calculus as an *object language* for reasoning mechanisms that support the secondary tasks. Arguments against Bayesian decision theory are weakened when the calculus is relegated to this role. Architectures for uncertainty handling that take statements in the calculus as objects to be reasoned *about* offer the prospect of retaining normative status with respect to decision making while supporting the other tasks in uncertain reasoning.

## 1 Introduction

Much of the controversy surrounding the use of uncertainty calculi in AI can be explained by a difference in the emphasis the competing approaches place on the various tasks faced by the problem-solver. In this essay, we argue that comparisons of competing uncertainty mechanisms in the context of any single task do not clarify the debate and further, that the attempt to completely describe the aspects of uncertainty necessary to support all of these tasks within a single "calculus" is misguided. Instead, we believe that the important advances in technology for building computer programs that reason under uncertainty will be achieved via knowledge representations that serve computational requirements normally considered beyond the realm of belief combination and decision making formalisms.[1]

*Supported by National Institutes of Health Grant No. R01 LM04493 from the National Library of Medicine. Address correspondence to the first author at MIT Laboratory for Computer Science, 545 Technology Square, Cambridge, MA 02139.

†Support for this work was provided by NASA-Ames and the National Library of Medicine under grant R01 LM04529. Computing facilities were provided by the SUMEX-AIM resource under NIH grant RR 00785.

[1] It is difficult to separate precisely the uncertainty "calculi" from "non-calculus" uncertainty representation mechanisms. Loosely speaking, we take calculi to be representations that directly attach some kind of "measure of belief" to proposition-like objects and provide a small set of combination rules for deriving belief measures for aggregations and transformations of the basic objects. Bayesian probabilities [12], Dempster-Shafer (D-S) belief functions [41], MYCIN certainty factors [45], and fuzzy possibility [54], for example, are well-known uncertainty representations that fall easily into the category of calculi.

We are driven to this view by the observation that knowledge engineering issues that arise in assembling and utilizing a knowledge base built on any uncertainty mechanism are not practically subject to the normative interpretations of calculus elements that may hold within the knowledge base. Such engineering concerns—perhaps not necessarily but always in practice—fall "outside the model." Therefore, arguments based only on properties of the calculus within the model cannot be conclusive.

These motivations resemble Shafer's well-stated arguments for "constructive probability" [42,44]. However, where Shafer concludes that constructive issues should guide comparison of calculi, we focus on the implications on representations and mechanisms supporting uncertain reasoning tasks.

## 2 Uncertain Reasoning Tasks

For concreteness, our analysis focuses on a mythical program for reasoning about medical problems, which we shall call the robot physician. Though we do not propose that technology of the near future will permit us to build a program comparable to humans in all respects, the demands of such a broad domain dictate strong design constraints on mechanisms for uncertain reasoning.

We start by putting ourselves in the place of a knowledge engineer embarking on the design of this robot physician. In the sections below, we discuss specific tasks that arise in uncertain reasoning to illustrate more forcefully the different representational issues that a knowledge engineer would face. We will see that no scheme yet presented is ideal for all of these tasks, and it is unrealistic to expect that any "framework" describable as an incremental extension to a calculus would be.

## 2.1 The One-Shot Decision

The basic task is one we call the "one-shot decision," which corresponds to the usual decision-theoretic formulation of a choice problem.

In the one-shot decision, the physician is required to choose from a set of alternate acts (for example, strategies involving combinations of diagnostic tests and drug therapies). To account for uncertainty, we postulate a state of nature that is unknown to the decision-maker (perhaps the identity and other features of the patient's disease) but is taken to be one of a set of possible states. The outcome of the decision (the resulting health and well-being of the patient) is determined by the actual state of nature and the chosen action.

A Bayesian decision-maker has a utility function over outcomes and a probability distribution over states of nature, and chooses the action that maximizes expected utility. In designing a robot physician for particular one-shot decisions, we can ensure that it is Bayesian by explicitly encoding its knowledge in the form of these objects—action sets, states of nature sets, utility functions, and probability distributions—using notations similar to those developed by decision analysts [29,37,40]. Indeed, given desiderata for belief measures and "rational decision making" [11,39], Bayesians can make a strong case for decision models as uniquely valid knowledge representations. Any weaker body of assertions may not justify a choice of action, and any additional information must be superfluous with respect to the decision.

Even those accepting the Bayesian arguments regarding the normative status of decision theory are free to reject the use of decision-theoretic models as a knowledge representation on pragmatic knowledge engineering grounds [50]. We consider this position in Section 3.

The pure one-shot decision scenario is an idealization that is only defensible in extremely time-constrained, highly-specialized situations in which a decision model must be pre-fabricated, perhaps in a hospital emergency room or during a surgical operation. More generally, a decision is embedded in a situation calling for additional reasoning capabilities.

## 2.2 Information Gathering

Once we relax the time constraint, our robot physician has several options in addition to the primary decision.

*Acquire more knowledge or information.* The robot physician may benefit from reading journal articles, exploring patient databases, or finding out more about the patient before making the decision in question. While information-gathering actions that possibly have serious medical consequences—invasive tests such as biopsies, for example—are often explicitly considered as part of the decision, relatively inexpensive steps like asking the patient questions are usually outside any formal decision model.

*Wait for uncertain events to resolve.* Sometimes valuable information can be obtained simply by waiting to see how things turn out. Because the benefits of a "wait" strategy are largely computational (we avoid planning for some contingencies that do not obtain anyway), incorporating this option within the primary decision model is technically difficult.

*Ask other physicians for advice.* The robot physician's colleagues (human or machine) may be better suited for this decision or might complement the robot's own expertise. The communication skills necessary for this strategy form an important class of reasoning tasks in their own right, and are discussed further in Section 2.3.

*Design and perform experiments.* In treating patients, information-gathering is typically limited to direct data acquisition or observing the results of pre-established experiments such as cell cultures. However, designing experiments (or at least somewhat modifying existing procedures) is always a real option to a decision-maker. For a robot research physician, experimentation would be a primary activity. To take advantage of this more active form of knowledge acquisition, the robot needs to understand the relationship between the uncertainty in its decision problem and other uncertainties that it can test in the world.

To perform any of these tasks—and even to select which of these avenues to pursue—our robot will need a very rich description of the state and nature of uncertainties involved in its decision environment. The knowledge prerequisites for effective information-gathering go well beyond the requirements for modeling the primary decision itself.

## 2.3 Communication

As mentioned above, one of the options open to our robot physician is to seek the advice of a human or machine colleague. Let us consider the communication tasks faced by two types of physician in this situation: a generalist and a specialist, either or both of which might be machines. For simplicity, we will assume the generalist is the decision maker or has direct access to the decision maker's preferences.

In a simple interaction, the generalist might consult a specialist dedicated to a particular decision problem. The model for the decision, including alternate acts, relevant events, outcomes, and probabilities, are assumed to be explicitly encoded in the knowledge base of this super-specialist. In such a situation, the generalist needs only communicate to the super-specialist the patient-specific details of the problem such as the patient's symptoms and preferences. The specialist then computes the optimal strategy using the calculus of decision theory and reports the result to the generalist. If an explanation is desired, the super-specialist could display to the generalist the sensitive portions of the model along with any needed discussion of decision theory.

In a more complex interaction between generalist and specialist, the generalist cannot assume that the specialist has a pre-existing decision model and therefore must present an accurate account of the decision problem to the expert. The consult-requesting generalist cannot in practice describe *all* of its information relevant to the problem; instead, it must choose appropriate levels of abstraction at which to convey its knowledge. For example, rather than presenting its consultant with a photographic image of the patient's retina, it might simply report that the fundascopic examination was "normal." In this communication task, the generalist must balance the costs of precise communication with the benefits of avoiding misinterpretation.

In reacting to this description, a specialist consultant must construct a model from the problem description and its "background knowledge." In doing so, it may identify or synthesize additional strategies not considered by the generalist, select state variables relevant to the problem, and characterize probabilistic relationships among the variables and strategies. Model construction issues are discussed further in Section 3.1.

Once the model is constructed, the robot specialist must present and possibly explain the implications of the model to the generalist. In many cases, a simple explanation like the one described above may be sufficient. Often, however, the generalist may not accept one or more components of the model adopted by the specialist. For example, the specialist's probabilities for uncertain events in the model may be different from those of the generalist and a discrepancy may still exist after the generalist hears the specialist state its beliefs. In this case, the specialist is faced with a complex explanation task: to identify and present pieces of knowledge to the consulting generalist that will convince it of the validity of the specialist's beliefs. We will refer to this complex explanation task as *justification*, recognizing that others have used this term to denote broader or somewhat different explanation activities.

Finding the optimal amount of knowledge and the level of abstraction at which to present it is a highly complex task. If too little or excessively abstract knowledge is conveyed, the physician will not be convinced. If too much or excessively fine-grained information is given, the physician will find the specialist a waste of time.

Most importantly, the information provided by the specialist should be a function of the knowledge or lack of knowledge of the agent seeking consultation. Knowledge valuable to one agent may be worthless to another.

This facet of justification has received some attention by artificial intelligence researchers under the name of user-modeling [46]. Knowledge representations consisting primarily of an uncertainty calculus (as employed by statistical diagnostic aids [1,25,49]) provide virtually no support for this task. While the volume of representation research in both uncertainty and epistemology has been great (enough to merit sizable conferences [20,27]), their intersection has been slight.

In discussing justification, we have assumed that the specialist's knowledge relevant to the decision problem is a superset of the generalist's knowledge. If this is not the case, the robot specialist or, more appropriately, the robot colleague, faces the task of trying to discover the information held by the consult requester that it does not currently possess. Similarly, the requester must try to pinpoint this information and present it to its colleague. In such a situation, our robots are faced with the extremely complex task of constructive debate.

## 2.4 Diversity of Reasoning Tasks: Discussion

Others have noted that different tasks may call for different uncertainty mechanisms [4]. In enumerating the reasoning tasks above, we emphasized the multi-plicity of requirements the various tasks impose on a representation scheme for uncertain knowledge. Because of the additional tasks surrounding any central choice problem, the decision model complete for the one-shot decision is no longer adequate. Competing knowledge representation mechanisms must be evaluated on their support for these other tasks. As we argue below, performance on these tasks is determined by structural and computational issues—not by selection of an underlying uncertainty calculus alone.

## 3 Knowledge Engineering Issues

The enumeration of tasks above demonstrated that the knowledge sufficient to make a one-shot decision does not necessarily support other, equally necessary, reasoning tasks. In this section, we present additional grounds for considering extra-calculitic issues even within the one-shot decision scenario. These arguments rule out the formulation of information-gathering and communication into a more general decision-making problem as an antidote for our task diversity conclusions.

A solution to the problem of building a robot physician must be sensitive to issues concerning the assembly of the robot's knowledge by its human designers. The knowledge representations that are best with respect to decision-making performance and computational efficiency may not be optimal or even feasible when it comes to knowledge engineering. The sections below provide several reasons that knowledge encoded in other forms might be preferable, and identify representational issues salient for computerized decision-making from non-decision-analytic knowledge bases.

### 3.1 Model Construction

In the discussion of reasoning tasks above (Section 2), we started with a representation oriented toward a primary decision and showed that this model failed to support the surrounding reasoning tasks. Here we start with a knowledge base not necessarily in the form of a decision model and consider the problem of formulating a model for the primary decision from this representation.

A decision model is suitable as a knowledge base only for the most specialized of robot physicians. A model covering more than a very narrow body of decision contexts is a poor one for any *particular* medical problem because the extraneous features considered tend to entail an unnecessary information-gathering burden and to obscure explanations of the result. General models cannot take advantage of simplifying features that—while present in any given decision problem—vary from case to case. In fact, we are aware of no decision-analytic models in medicine applicable outside of a narrowly-defined class of patient cases. This limitation is not restricted to decision analysis technology, however; AI-style medical expert systems that generate treatment recommendations have invariably been super-specialists.[2]

One approach toward overcoming the apparent unscalability of decision models is to build a knowledge base in some other form and endeavor to customize a decision model for each given problem instance. Unfortunately, we have to be rather vague about these "other forms" of knowledge representation because efforts to build large knowledge bases in AI projects to date have met with only mixed success. Nevertheless, we cite below several features of AI techniques which can undoubtedly enhance the extensibility of existing computational mechanisms for decision modeling.

Broadly speaking, decision models and therefore the task of constructing them can be partitioned into three components:

- generating possible strategies,
- identifying relevant variables, and
- assignment of beliefs and preferences.

The target representation for a decision-model constructor must have objects corresponding to the strategies, events, probabilities, and utilities produced by these component tasks. Note that the decision-model formalisms cited above may be suitable target representations despite their inadequacy for general knowledge representation in the robot.

The potential advantages of alternate knowledge representations lie in the possibility of implicitly capturing an enormous variety of decision models in a relatively compact encoding. For example, in traditional AI planning,[3] the set of strategies is implicitly represented by a description of the action types and a set of combination rules. Thus, we can specify the strategies open to our robot physician by describing the individual actions that it can prescribe (various drugs and their dosage ranges, diagnostic tests, and surgical procedures, for example) along with an account of how they can be sequenced, conditionalized, or otherwise combined into complete therapy plans. ONYX, a planning architecture developed by Langlotz et al. [30], combines knowledge represented in this form with patient-specific data to generate a reasonable number of plans to consider for a particular case. The best strategy is then identified via a decision-analytic evaluation model. In contrast, a direct decision model representation would have to enumerate the complete therapy plans in advance. Outside of extraordinarily narrow decision contexts, supplying such an exhaustive list would not be feasible.

---

[2]To verify these assertions, see Kassirer et al. [28] for a review of medical decision analysis applications. The papers collected by Clancey and Shortliffe [8] provide a perspective on the state-of-the-art of decision-making in knowledge-based systems for medicine. Rennels et al. [38] analyze some of the choice mechanisms applied in these systems.

[3]See Charniak and McDermott for an overview of AI work in planning [5]. Wellman [52] recasts the AI planning paradigm in a framework compatible with decision theory.

Likewise, the set of possible events to include in the target decision model may be combinatorial in some more primitive type of element, for example if events can be described as patterns of sub-events over time. Or perhaps variations on events are describable by combinations of lower-level features. If so, a terminological knowledge representation facility (exemplified by KL-ONE [3]) would allow the knowledge engineer to describe structural relations among features, leaving instantiation of precise feature combinations (an exponential number of possibilities) to the requirements of a particular problem instance. In either case, the static decision model form of knowledge representation fails to take advantage of structural regularity because it requires that the most detailed descriptions of events explicitly appear at the surface of the knowledge base.

Many of these advantages may be achieved at least in part by incremental enhancements to existing decision-model computational tools. Regardless of implementation perspective, the knowledge engineering benefits accrue from the adoption of knowledge representations separate from the decision models produced for the final analysis.

### 3.2 Knowledge Modification

An important concern for the representation of knowledge is the ease with which a representation can be modified. In real-world applications, a knowledge base that is difficult to modify soon becomes obsolete. The key characteristic of a representation when it comes to modifying the knowledge base is modularity. Modularity is an engineering concern for any knowledge-based system; under uncertainty the problem is magnified by the sensitivity of probabilistic relations to surrounding context [23].

To illustrate the modularity problems that arise in probabilistic knowledge representations, we adduce an example from Cooper's NESTOR [10] that was used by Spiegelhalter [48] for another purpose. NESTOR's domain is hypercalcemia, hence the program includes a knowledge base relating a patient's calcium level to other physiological states and associated findings. In Spiegelhalter's model fragment, the unconditional or prior probability of coma is .05. To us non-specialists this value seems high; a much lower fraction of people we know are comatose. Of course, the number may be valid for the population treated by the program: patients identified somehow for a hypercalcemia work-up.

The point is that the knowledge base contains no definition for the population it is applicable to. Spiegelhalter and Knill-Jones [49] discuss the issue of transportability of statistical knowledge bases, concluding that it remains a serious problem. A greater difficulty, in our view, is the limitation it imposes on the scope of any such knowledge base. Suppose we wished to extend NESTOR's domain by including medical knowledge from another program working in a neighboring or overlapping clinical area. Patil's ABEL [35], for example, models disorders of electrolytes other than calcium, and considers more deeply the acid/base issues relevant to hypercalcemia. In broadening the domain, we have no idea which parts of the hypercalcemia model remain valid and which must be changed in light of the modified population and new interacting variables. Although the same is strictly true of ABEL's knowledge—we cannot be certain that the electrolyte model is still correct when calcium is considered explicitly—the causal link structure of ABEL is more robust than the precise statistical relationships.

The robustness of causal links rests partly in their imprecision (weaker statements hold in more contexts and are therefore more modular), but also in that they capture a critical aspect of the domain knowledge. A causal model with a surface representation reflecting a theory or set of organizing principles underlying the domain should be less sensitive to context than an arbitrary selection of observed relationships among variables.

### 3.3 Multiple Knowledge Sources

The majority of knowledge-based computer programs are modeled not on single human experts but on expertise gleaned from multiple sources. A ver-satile robot physician should (like its human counterpart) include knowledge based on several human experts, as well as medical texts, journal articles, databases, and observed clinical results.

Unfortunately, a knowledge engineer has no available standard for combining knowledge from these disparate sources into a single coherent belief model. In fact, even if each of the individual sources were stated in terms of consistent Bayesian decision models, there is no accepted normative procedure for deriving a Bayesian model from their combination [15]. Furthermore, there is little reason to believe that a purely calculus-based algorithm could be developed because an effective combination procedure would need to take into account the strengths and weaknesses of each knowledge source, as well as the nature of disagreements among the sources.

## 4 Decision Calculus as Object Language

Some may still object that our separation of the secondary tasks from the primary decision is artificial. For example, the problem of the colleague presenting knowledge to the physician can be framed in decision-theoretic terms: select the fraction, level of abstraction, and presentation of the knowledge that would be of maximum utility to the physician. Theoretically, it is possible to reformulate each task as a decision problem, here by treating communication acts as decision options. However, we find this solution unsatisfactory for two main reasons:

- Viewing these tasks as decision problems fails to take advantage of their special structure.

- The resulting decision problem overwhelms our practical model-building capabilities.

This latter point summarizes the lesson we draw from our analysis of knowledge engineering issues in Section 3.

An approach we have implicitly advocated elsewhere in this paper takes the uncertainty calculus as part of an object language to be reasoned *about* by the decision-making computer program. Assertions stated in terms of calculus elements (in particular, probabilities) are derived from the knowledge base, which need not encode facts directly in that form. To choose a plan of action, the program *constructs* a decision model founded on the formal calculus. Calculus-based representations may or may not play a role in structures employed to support reasoning tasks surrounding the primary decision, for example, user models built to aid communication.

This position has been expressed by previous AI researchers. Even in pronouncing probability "epistemologically inadequate," McCarthy and Hayes acknowledge that "the formalism will eventually have to allow statements *about* the probabilities of events" (emphasis added) [33, page 490]. More recently, Grosof has provided a first-order-logic formulation treating probabilistic statements as terms within a meta-language [18,19].

In the remainder of this paper we develop this solution approach, taking the object language to be Bayesian decision theory. As suggested in Section 2.1, once the calculus is restricted to this role the case for decision-theoretic semantics becomes compelling. Motivations for introducing variant calculi are reduced because secondary tasks and knowledge engineering issues are addressed by extra-calculitic mechanisms. In this context, general arguments for probability [6,24] as well as specific criticisms of alternate calculi (MYCIN certainty factors [22] and fuzzy probability [7,14], for example) are difficult to refute.

## 5 Incomplete Decision Models

A direct implementation of the decision-calculus-as-object-language approach may be feasible—if we relax the strict requirement that decision models be complete in the Bayesian sense of full specification of probabilities and utilities.[4] Indeed, many uncertainty mechanisms offered in the literature highlight incompleteness as a feature: D-S belief functions [41], Cooper's NESTOR [10], and Good's lower probabilities [16], to name but a few. While incompleteness is not a panacea, we will see that allowing partial models may offer substantial advantages with respect to each of the reasoning tasks and knowledge engineering issues described above.

### 5.1 Incompleteness and Evidential Structure

One of the most-cited virtues of uncertainty calculi that admit incompleteness is that they allow the rep-resentation to express ignorance about the propositions in question. Although in Section 5.2 we dispute the claim that incompleteness straightforwardly captures ignorance, allowing decision models to be incomplete does facilitate the expression of evidential structure and model derivation (that is, source of beliefs)—necessary preludes to any representation of "ignorance." Ignorance aside, these are precisely the sorts of knowledge irrelevant for one-shot decisions yet indispensable for the surrounding reasoning tasks.

---

[4] "Completeness" as used here is only defined with respect to a particular space of designated events. A model complete for a coarse-grained event space may be incomplete for a more refined space or for one defined by overlapping propositions. The incompleteness we are interested in permits the problem-solver to refer to events for which probability distributions are unavailable.

That belief functions offer advantages in representing evidential structure is the keystone of the case for the Dempster-Shafer uncertainty calculus. In pleading the case, Shafer relates a model for evidence acquisition as receiving a sequence of noisy messages asserting that the true state of nature lies in various hypothesis sets [43]. For some environments, this model appears to correspond well to actual information-gathering processes. In the application of Lowrance et al. [32], for example, information input is of the form "the ship location is in the set $A$." Other applications of the D-S calculus stretch the interpretation of Shafer's evidence model somewhat beyond its elasticity. Lemmer [31] argues that the use of statistically derived data is inconsistent with this formulation, thereby invalidating numerous applications in the literature.

A belief function represents evidence structure by distinguishing knowledge based on a single piece of evidence supporting $A = \{a_1, \ldots, a_n\}$ from an equivalent (in the Bayesian sense) belief state based on $n$ weaker pieces of evidence supporting the individual $a_i$s. But arguments that this is a good representation for evidential structure must ultimately be settled on empirical grounds, as Shafer points out [42, page 15]. Connections of evidence to sets of hypotheses is only one component of evidential structure—other features of evidence patterns not captured in the D-S belief function decomposition would surely be useful as well. For example, incompleteness arising from logical combinations of propositions in Nilsson's probabilistic logic [34] might be viewed as a representation of the "boolean pattern" of evidence. There is no *a priori* reason to think that a single uncertainty calculus—D-S belief functions or one yet to be invented—can provide a universally appropriate means for expressing evidential structure.

The issue of evidential structure representation can and should be divorced from degree-of-belief aspects of a calculus. Evidence structuring mechanisms inspired by the D-S approach are generally implementable within other formalisms. For example, Pearl [36] was able to achieve many of the benefits of Gordon and Shortliffe's hierarchical D-S scheme [17] with a Bayesian mechanism.[5] And other models of evidence may be better suited for a Bayesian treatment than one based on belief functions.

Currently, there is little empirical support for the effectiveness of D-S or any other particular mechanism for representing evidential structure. We also lack, for that matter, strong demonstrations of the benefits of evidence structure representation for uncertain reasoning tasks. But despite the lamentable dearth of concrete experience, an examination of the tasks themselves suggests that evidence structure is one of the components of a good uncertainty representation, and that incompleteness can play a supporting role in capturing this structure.

### 5.2 Incompleteness, Reasoning Tasks, and Knowledge Engineering

A close look at the reasoning tasks and knowledge engineering issues discussed in Sections 2 and 3 reveals that many of the reasoning problems identified are aggravated substantially by the requirement of completeness for decision models. Some brief examples:

- *Information Gathering.* The prospect of receiving additional data, facts, and advice pertaining to the decision problem entails major expansions to any complete Bayesian belief model. Although structural regularities such as conditional independence conditions may be exploited to reduce the assessment task, construction of complete models for all but the most narrow decision contexts is likely to be intractable.

- *Communication.* The common approaches to explanation exploit the structure of evidence leading the program to its conclusions. If the uncertainty representations that best capture evidential structure happen to be incomplete decision models, insisting on completeness degrades the program's explanatory ability.

- *Model Construction.* Model construction from a modular knowledge base is largely an assembly

---

[5] Pearl interprets the relation of evidence to a set of hypotheses $S$ as an assertion of conditional independence between the evidence and subsets of $S$ given $S$. Because his algorithm performs complete propagation based on this assertion, the representation does not explicitly maintain incompleteness and therefore cannot reconstruct the evidence pattern. However, as Pearl notes, the propagation steps could be postponed with appropriate bookkeeping, thereby providing the desired evidence feature.

task: piecing together a decision model from distributed fragments. In this decentralized framework, it is difficult to ensure that the resulting assembly is complete and consistent. A collection of submodels, each complete in isolation may underspecify the situation in combination.

- *Knowledge Modification* Incomplete models possess a modularity advantage over complete models for the simple reason that weak statements hold in a wider variety of contexts than strong ones.

- *Multiple Knowledge Sources.* As noted above, when a knowledge base is derived from a group of experts or a combination of sources, there is no accepted normative procedure for integrating the opinions and judgments into a single Bayesian decision model [15].

Two main points arise from our survey of incompleteness and its relation to evidential structure, reasoning tasks, and knowledge engineering.

First, the benefits of permitting incompleteness in decision models are substantial across the board. Though these benefits are intangible (or at least difficult to quantify within the decision formalism itself) and subject to debate, they cannot be dismissed as negligible or irrelevant. In the sense that any model can only approximate reality [47], true completeness is an unattainable standard anyway. Those concerned with indecisiveness resulting from incomplete models should be persuaded to examine the tradeoff more closely, as we shall in Section 6.

Second, the diversity of uses for incompleteness suggests that a calculus oriented toward a single form of incomplete model cannot be ideal. Instead, the calculus is better treated as a decision model object language with a variety of structural constructs for expressing the incompleteness. It is reasonable to expect that different representation schemes will be best suited to the various tasks.

A corollary to this second point is that simple interpretations of the amount of incompleteness cannot be correct. Without knowing the reason for the lack of full model specification (modularity considerations, limited modeling resources, or other possibilities suggested above), one cannot maintain that a measure of incompleteness expresses "ignorance" or any other particular quality. Ignorance and incompleteness are two separate issues and should be strictly distinguished. Failure to do so is a prime source of common misconceptions about uncertainty calculi.

# 6 Making Decisions from Incomplete Models

In allowing incompleteness we sacrifice guaranteed decisiveness in our decision models. Many approaches to decision making in underdetermined situations have been proposed over the years; we briefly review some of them in this section. Our intent is to provide a sense of the options available rather than to endorse a particular technique.

### Complete the Model

One category of techniques comprises those that transform the incomplete model into a complete one by filling in the gaps. The method receiving the most theoretical and practical attention is to choose the complete probability model maximizing entropy among those consistent with the incomplete model provided [26]. Grosof [19] demonstrates that this and other "gap-filling" approaches are forms of nonmonotonic reasoning about statements in the uncertainty calculus.

### Heuristic Decision Rules

Methods within a second category generate decisions directly from incomplete models. Though lacking normative justification, these techniques can be computational bargains. The prototypical decision rule for incomplete models is the minimax loss criterion studied in statistics [2]. Methods based on structural properties of the evidence and arguments salient to the decision, such as Cohen's endorsement theory [9], are also examples of this category.

### Further Computation and Assessment

The third approach is based on a premise that further computation and/or assessment can eventually achieve decisiveness. To converge on a decision, computation may be directed toward either refinement of the decision model or dominance-proving activities. Decision model refinement falls within the model construction task we discussed in Section 3.1. Dominance-proving involves computational manipulations of the incomplete decision model aimed at defining the set of admissible strategies.

Little research effort has been expended to date on architectures and languages for general-purpose decision proving, although numerous techniques applicable to special cases have been developed [13, 21,51,53]. The variety of dominance-proving strategies available seems to indicate that several different forms of incompleteness must be accommodated; undoubtedly many of these will be unlike the kinds of incompleteness useful for capturing evidential structure.

**Decision Making: Discussion**

The far-from-exhaustive review above suggests that a variety of remedies are available for indecisiveness induced by incompleteness in decision models. All are oriented towards quantitative incompleteness; missing strategies, events, or other structural model features are not addressed.

In addition, each of the methods has its own limitations pertaining to the forms of incomplete models they accommodate and the reasons for incompleteness on which they are justified. Comparisons among them must be performed on empirical grounds, or theoretically within a specific framework for incompleteness. The best decision-making policy for any practical program probably consists of a mixture of the above and other approaches. Assessment of the costs and benefits of each method for the types of incomplete models they are to be applied to should be used to determine which method or combination is most appropriate for the particular task at hand.

# 7 Conclusion

Our observations over the preceding sections lead us to a series of methodological conclusions for the development of uncertain reasoning mechanisms.

1. Uncertainty representations sufficient for one-shot decision-making may be inadequate for the larger reasoning problem that encompasses supporting tasks such as information-gathering and communication.

2. The variety of knowledge engineering concerns makes it unlikely that incremental extensions to an existing calculus will offer benefits for a significant fraction of them.

3. Approaches that reason about the elements of a calculus place the burden of supporting the various reasoning tasks on *extra-calculitic* mechanisms. Once the responsibility is thus transferred, the case for a calculus derived from Bayesian decision theory is considerably strengthened.

4. Numerous engineering objectives are served by permitting incomplete decision models in the underlying representation. However, this multiplicity of objectives also requires that many forms and interpretations of incompleteness be accommodated. Again, no "calculus" will suffice.

5. Although the possibility of indecisiveness is an unavoidable by-product of incompleteness, many computational courses are open to programs faced with such situations. Further research in this area appears promising.

In this essay, we have tried to unify a growing body of research that implicitly or explicitly treats the decision calculus as an object language manipulated by higher-order mechanisms for uncertain reasoning. Those who would invent calculi to directly address some of the knowledge engineering issues are asking too much from such a limited class of representation mechanisms. A more reasonable role for uncertainty calculi in knowledge-based computer programs is to provide a grounding in decision theory, thereby offering a semblance of normative status to the decisions made by these programs.

**Acknowledgments**

Jack Breese and an anonymous referee provided valuable suggestions for this paper. We have also benefitted from discussions with Peter Cheeseman, Greg Cooper, Ben Grosof, Eric Horvitz, Curt Langlotz, and Peter Szolovits.